\documentclass{article}
\usepackage{ijcai26}

\usepackage{times}
\usepackage{soul}
\usepackage{url}
\usepackage[hidelinks]{hyperref}
\hypersetup{
    pdftitle={Robust Contrastive Graph Clustering with Adaptive Local-Global Integration},
    pdfauthor={Lei Zhang, Fubo Sun, Haipeng Yang, Zhong Guan, Likang Wu},
    pdfsubject={Graph Clustering},
    pdfkeywords={Graph Neural Networks, Contrastive Learning}
}

\usepackage[utf8]{inputenc}
\usepackage[small]{caption}
\usepackage{graphicx}
\usepackage{amsmath}
\usepackage{amsthm}
\usepackage{booktabs}
\usepackage{algorithm}
\usepackage{algorithmic}
\usepackage[switch]{lineno}

\usepackage{graphicx}
\usepackage{amsthm,amsmath,amssymb,multirow, amsfonts,subfigure}
\usepackage{wrapfig}
\usepackage{xcolor}
\usepackage{subcaption}
\usepackage[switch]{lineno}    
\usepackage{afterpage}

\title{Robust Contrastive Graph Clustering with \\Adaptive Local-Global Integration}

\author{
Lei	Zhang$^1$
\and
Fubo	Sun$^1$\and
Haipeng	Yang$^1$\and
Zhong	Guan$^2$\And
Likang	Wu$^2$\thanks{Corresponding Author.}\\
\affiliations
$^1$School of Computer Science and Technology, Anhui University\\
$^2$College of Management and Economics, Tianjin University\\
\emails
zl@ahu.edu.cn,
e24201024@stu.ahu.edu.cn,
haipengyang@126.com,
kerkersit21@gmail.com,
wulk@tju.edu.cn
}

\begin{document}

\maketitle
\begin{abstract}
Graph clustering is essential in graph analysis for revealing structural patterns and node communities. Despite recent advances in self-supervised contrastive learning that have improved clustering via structural and attribute signals, existing methods still struggle to flexibly capture high-order local structures and often overlook global semantics in complex graphs. These limitations lead to suboptimal node representations, especially in real-world graphs with fragmented structures and ambiguous cluster boundaries. To address these limitations, a contrastive graph clustering framework is proposed to jointly integrate multi-scale local structures with global semantics via attention mechanisms. At the local level, GNN-based topological signals extracted from multiple propagation depths are adaptively fused through attention-based weighting to capture multi-scale neighborhood features. At the global level, semantic prototypes derived from dynamically evolving cluster centers are adaptively aggregated through attention to guide node representations and enhance inter-cluster separability. The model is trained under a dual-view contrastive learning paradigm with a hybrid objective that combines instance-level and structure-aware losses to improve representation robustness and discrimination. Experiments on eight real-world graph datasets demonstrate that our method achieves competitive clustering performance. Code is  available at \url{https://github.com/vege12138/w2}.
\end{abstract}

\section{Introduction}
Graph clustering is a fundamental task in graph analysis ever, aiming to partition nodes into clusters or communities based on structural attributes or connectivity patterns, which is essential for uncovering underlying patterns and relationships among nodes. This task has been studied and applied in domains including social network analysis~\cite{qiu2018deepinf}, knowledge graphs~\cite{BELLOMARINI2022101528}, and recommender systems~\cite{fan2019graph,DBLP:conf/www/WuLCLCNJJS025}. As a long-standing research topic, graph clustering continues to evolve.

To advance deep graph clustering, Graph Convolutional Networks (GNNs)~\cite{kipf2016semi} have been employed to learn node representations through message passing. Building on this, self-supervised graph representation learning has emerged as a promising paradigm, leveraging intrinsic structural and attribute information as supervisory signals to guide similar nodes toward compact embeddings. Existing approaches typically follow generative or contrastive paradigms. Generative methods reconstruct graph topology and features via an encoder–decoder architecture, with enhancements combining MLPs and GCNs to capture semantic and structural cues~\cite{tu2021deep}. In contrast, contrastive learning constructs positive and negative sample pairs to maximize mutual information between positives while minimizing it for negatives~\cite{peng2024multi}.
\begin{figure}[t]
    \centering
    \includegraphics[width=0.46\textwidth]{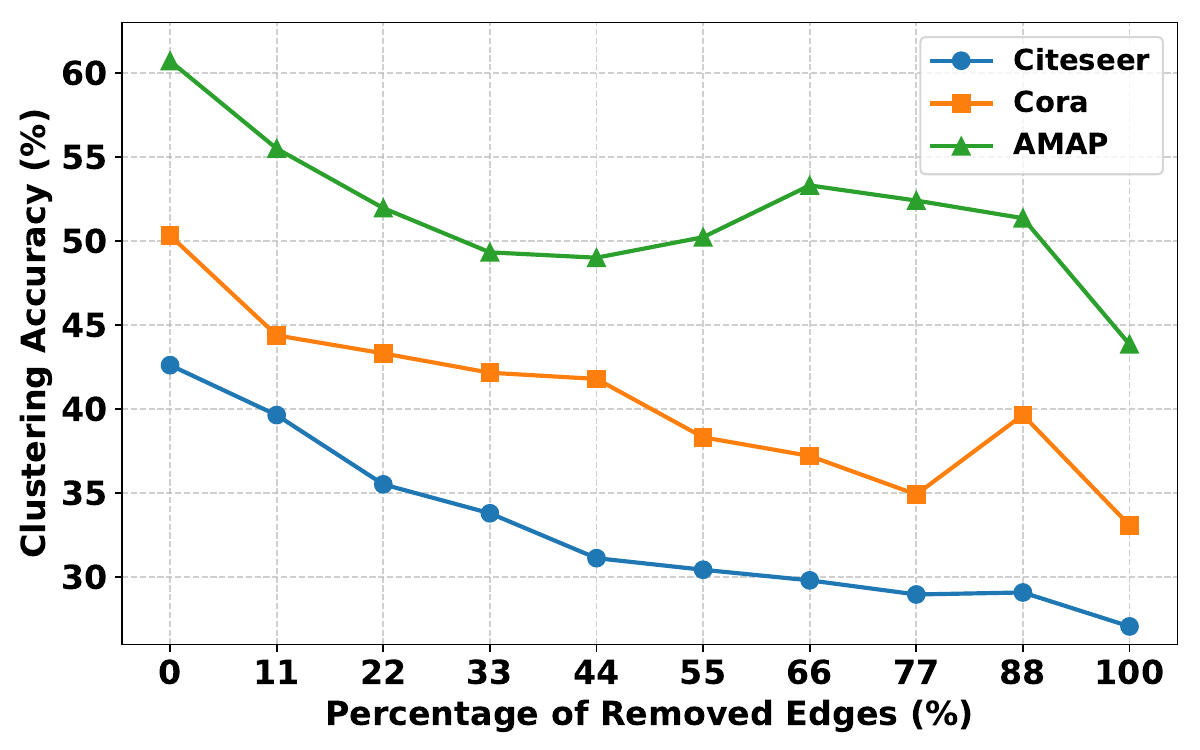}
    \caption{Clustering accuracy (\%) versus the percentage of removed high betweenness centrality edges (up to 2700 edges) on three datasets: Cora, Citeseer, and AMAP.}
    \label{fig:acc_vs_components}
\end{figure}

Despite significant advances in deep graph clustering, most existing methods focus primarily on local neighborhood aggregation while overlooking the integration of high-order local structures and global context. This often results in unstable and suboptimal node embeddings, especially in real-world graphs where cluster boundaries are ambiguous. Nodes situated near community intersections or in fragmented subgraphs face particular challenges: they require adaptive fusion of multi-scale neighborhood information across shallow and deep receptive fields, but conventional message passing either ignores this heterogeneity or introduces noise from distant nodes. 
Moreover, many real-world graphs are fragmented, leaving isolated or weakly connected components with limited access to high-level contextual signals and making cross-community discrimination unreliable. Without global awareness, long-range dependencies become fragile and embeddings for such nodes may collapse. 
In this work, a basic clustering pipeline is adopted in which GNN-based encoders produce node embeddings that are clustered by $k$-means; however, as high-betweenness edges are removed, clustering performance further deteriorates (see Figure~\ref{fig:acc_vs_components}).
 
To address these limitations, this paper proposes a novel graph clustering framework that jointly integrates multi-scale local structural information and global semantic prototypes in an adaptive manner. At the local level, node representations with different receptive fields are fused through a purpose-built multi-head attention mechanism rather than a generic combination of GNN and attention. Specifically, features from different propagation layers are aggregated via max-pooling and scaled dot-product attention, enabling each node to selectively absorb relevant shallow or deep neighborhood cues. This design directly alleviates the over-smoothing and noise accumulation problems in rigid message-passing schemes, while improving the ability to capture hierarchical structural patterns. At the global level, dynamically updated cluster centers act as semantic prototypes. Through a cluster-guided attention mechanism, nodes selectively integrate global signals that evolve with the latent clustering structure. Unlike static global aggregation, this dynamic strategy adapts to structural changes and is particularly beneficial for nodes in sparse or fragmented regions, addressing the unreliability of long-range dependencies in real-world graphs. This dual-level integration enables the model to handle both local structural ambiguity and global fragmentation.

In addition, the framework adopts a contrastive learning paradigm by generating two augmented graph views to enable label-free representation learning. This is particularly important in unsupervised graph clustering, where the absence of ground-truth labels often leads to collapsed node representations. By enforcing consistency between different views of the same node (via instance-level contrastive loss) and leveraging graph topology to align structural neighbors (via neighbor-supervised loss), the model mitigates collapse and enhances structural discrimination. Consequently, this hybrid training objective yields embeddings that are more robust and consistent, and generalize better across diverse graphs. The main contributions of this work are summarized as follows:

\begin{itemize}
    \item We propose RCLG, a robust contrastive graph clustering framework that integrates multi-scale local structural cues with global semantic context via attention mechanisms under a dual-view self-supervised paradigm.

    \item We introduce two complementary components for local-global integration: a multi-head attention-based local fusion module that adaptively weights multi-depth representations, and a prototype-guided global aggregation module that uses dynamic cluster centers to inject global context and improve inter-cluster separability in sparse or fragmented graphs.

    \item The proposed model demonstrates strong robustness and stability. Without the need for dataset-specific hyperparameter tuning, it achieves consistently competitive clustering performance across diverse graph datasets with varying structures, highlighting its practical utility and deployment-friendliness.
\end{itemize}

\section{Related Work}
This section reviews the literature from two key perspectives.
The first part surveys the landscape of \textit{Deep Graph Clustering}, providing foundational context for this research.
The second part examines a key challenge in this domain: effective capture of \textit{Long-range Dependencies in Graphs}, analyzing the capabilities and limitations of existing approaches.

\subsection{Deep Graph Clustering}

Deep graph clustering aims to partition graph nodes into meaningful communities by leveraging deep neural networks to jointly capture both node attributes and structural dependencies~\cite{DBLP:conf/www/ZhaoLZWZL25}. With the introduction of GNNs, this task has witnessed significant progress, as GNNs offer a natural way to propagate and fuse feature information over graph topology. By extending the principles of deep clustering, graph-based approaches no longer treat each sample in isolation but instead emphasize the relational structure among nodes, enabling more coherent clustering decisions.

Graph contrastive learning~(GCL) \cite{xu2021infogcl,xia2022simgrace} is a self-supervised learning method that has been widely studied in various research areas~\cite{zhang2022enhancing,DBLP:conf/kdd/ZhengZLCCZW25}, such as node classification, recommendation, and clustering. Inspired by the success of GCL in self-supervised representation learning, several works have adopted GCL-based techniques to enhance cluster separability in an unsupervised manner~\cite{S3GC,DBLP:conf/mm/MaZ23,CCGC,peng2025sola}. Models such as SCGC and RGC initiate clustering by employing shallow graph convolutions to encode local interactions, followed by contrastive learning to capture relational patterns among samples. However, their shallow architectures limit the expressiveness of node embeddings, especially in capturing complex dependencies, which hinders clustering performance.

To address this, methods like SDCN~\cite{bo2020structural} and AGCN~\cite{peng2021attention} introduce deeper GNN architectures and attribute-structure fusion mechanisms to refine representation quality. These approaches demonstrate that deeper propagation can enrich the semantic content of node embeddings and better capture latent cluster structures. Nevertheless, they still suffer from limitations such as rigid neighborhood aggregation and inadequate handling of higher-order local information. Specifically, they often apply uniform message passing schemes across the graph, which can lead to noise accumulation or oversmoothing, particularly for nodes near community boundaries or in fragmented subgraphs.

\begin{figure*}[t]
    \centering
    \includegraphics[width=0.85\textwidth]{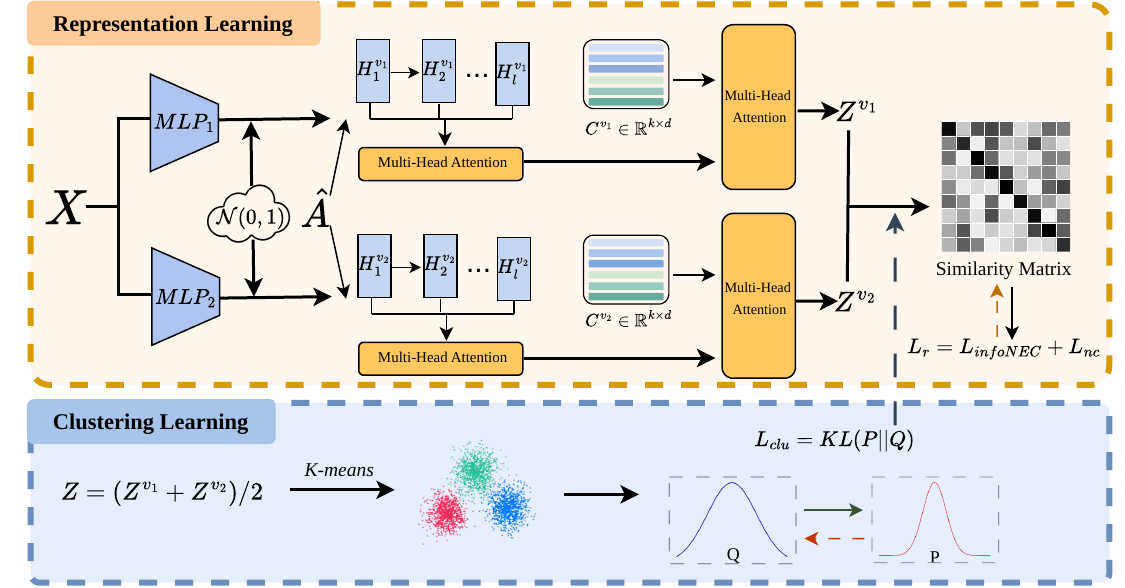}
    \caption{Architecture of the RCLG Model. Within the contrastive learning framework, the model first encodes node features \(X\) with noise perturbations. Subsequently, it employs multi-head attention mechanisms to integrate local information through the combination matrix \(\hat{A}\) and global information through the cluster centers \(C^{v_e}\). The model then utilizes both representation learning loss and clustering loss $L_r$ to achieve graph clustering. }
    \label{fig:framework}

\end{figure*}
\subsection{Long-range Dependencies in Graphs}
Capturing long-range dependencies is critical in graph learning, particularly for clustering tasks where nodes within the same community may be distant in the graph topology. Early methods, such as JKNet~\cite{xu2018representation} and DAGNN~\cite{liu2020towards}, attempt to aggregate information using layer-wise skip connections or adaptive weighted mechanisms to capture local features. BM-GCN~\cite{he2022block} and HOG-GCN~\cite{wang2022powerful} consider all neighbors to aggregate global information. However, these methods rely on the original graph topology, which makes them vulnerable to the influence of structural noise or sparsity in the graph.

Recently, MGCN~\cite{li2024multi} demonstrates the effectiveness of adaptively fusing multi-layer receptive field information for graph clustering. By incorporating deeper neighborhood contexts, it enables the aggregation of higher-order local structural information, beneficial for capturing nuanced intra-cluster relationships. However, these approaches typically rely on the assumption that the input graph is well-connected. In real-world scenarios, graphs are often fragmented, exhibiting isolated or weakly connected components. In such fragmented structures, traditional message-passing mechanisms either fail to transmit meaningful global signals or suffer from excessive noise introduced by irrelevant paths. As a result, embeddings of nodes—especially those in sparse regions—become unstable and less discriminative. Although DGAC~\cite{KunDGAC2025} is capable of capturing high-order semantic information, it can only marginally handle the issue of missing edges in the original graph. Therefore, effective local integration of higher-order structural information and strong global semantic guidance are both crucial for robust clustering, especially under sparse or fragmented graph connectivity.

\section{Methodology}
\subsection{Overview}
We use the following notations throughout this paper. An undirected graph is defined as $G = (V, E)$, where $V = \{v_1, v_2, \dots, v_N\}$ denotes the set of $N$ nodes and $E \subseteq V \times V$ is the edge set. The node features are represented by the matrix $X \in \mathbb{R}^{N \times d}$, where $d$ is the input feature dimension.
The structure of the graph is described by the adjacency matrix $A \in \{0, 1\}^{N \times N}$, where $A_{ij} = 1$ if an edge exists between nodes $v_i$ and $v_j$, and $A_{ij} = 0$ otherwise. The degree matrix is denoted as $D \in \mathbb{R}^{N \times N}$, where each diagonal element $D_{ii}$ equals the degree of node $v_i$. Incorporating self-loops involves defining $\tilde{A} = A + I$ and $\tilde{D} = D + I$, with $I$ representing the identity matrix. Given an unlabeled graph $G=(V,E)$ with node features $X$, graph clustering aims to partition the nodes into $K$ disjoint clusters, denoted by an assignment vector $y \in \{1,\dots,K\}^{N}$, such that nodes within the same cluster are highly similar while nodes from different clusters are well separated in the learned representation space.

In this section, the proposed model framework for addressing the above problem is introduced.  As shown in Figure~\ref{fig:framework}, the method adopts a contrastive learning framework composed of two structurally identical views. For each view, input node features are first encoded, followed by local and global information integration through multi-head attention. This design captures multi-scale structural information and enhances representation robustness. Unless otherwise specified, the following subsections describe the node representation learning process from the perspective of a single view, denoted as view \(e\) (where \(e = 1\) or \(2\)). 
In what follows, the framework is introduced component by component starting from view construction at the input stage, then detailing local and global fusion modules, and finally presenting the overall training objective.

\subsection{Noise-based View Augmentation for Robust Contrastive Learning}

To improve the robustness of node representations and enhance contrastive learning, Gaussian noise is injected during feature encoding. Given the input node feature matrix $X$, view-specific embeddings are generated by applying an MLP followed by noise addition:

\begin{equation}
Z^{v_e} = \text{MLP}_e(X) + \alpha \cdot \varepsilon, \quad \varepsilon \sim \mathcal{N}(0, I),
\end{equation}
where $Z^{v_e} \in \mathbb{R}^{N \times d}$ denotes the embeddings for view $e$, $\alpha$ controls the noise level, and $\varepsilon$ is standard Gaussian noise. This formulation preserves structural consistency while introducing feature perturbations, promoting harder positives and preventing collapse.

Unlike graph-specific augmentations such as edge dropping or diffusion—which are often complex and dataset-sensitive—this feature-space augmentation is simple, efficient, and broadly applicable~\cite{liu2023simple}. It offers a lightweight yet effective means of generating distinct but semantically consistent views, facilitating robust contrastive learning in subsequent fusion stages.

\subsection{Multi-level Local Feature Fusion via Attention}

To capture multi-scale local structural information, repeated message passing is performed over the noisy node embeddings $Z^{v_e}$ using the normalized adjacency matrix $\hat{A}$. Specifically, for each propagation step $t$, the corresponding layer-wise representation $H_t$ is computed, and all such representations are stacked to obtain a multi-layer feature tensor $S$:

\begin{equation} 
\begin{aligned}
& H_t = \hat{A}^t Z^{v_e}, \quad t = 1, 2, \ldots, l
\\
& S = \text{stack}(H_1, H_2, \ldots, H_l) \in \mathbb{R}^{N \times l \times d},
\end{aligned}
\end{equation}
where \( \hat{A} = \tilde{D}^{-\frac{1}{2}} \tilde{A} \tilde{D}^{-\frac{1}{2}} \) is the symmetrically normalized adjacency matrix, $Z^{v_e} \in \mathbb{R}^{N \times d}$ denotes the initial node embeddings in view $v_e$, $H_t$ is the node representation after $t$-hop propagation, and $S$ is the resulting stacked tensor comprising multi-scale representations for each node.

To adaptively select the most informative layer-wise features for each node, max-pooling is applied across all $H_t$, followed by a linear transformation to obtain the query matrix for attention weighting.

\begin{equation}
Q_L = \left( \max_{1 \leq t \leq l} H_t \right) W^q_L \in \mathbb{R}^{N \times d},
\end{equation}
where \( \left( \max_{1 \leq t \leq l} H_t \right) \) denotes the maximum across all layers for each dimension, and $W^q_L \in \mathbb{R}^{d \times d}$ is a learnable matrix transforming pooled features into query vectors for attention.

A scaled dot-product attention mechanism is employed to compute the attention weights between $Q_L$ and each layer in $S$, and the resulting scores are used to aggregate the layer-wise representations. A residual connection and layer normalization are applied to obtain the locally fused embeddings $Z^{v_e}_L$:

\begin{equation}
\begin{aligned}
& Z^{v_e}_L = \text{softmax} \left( \frac{Q_L S^\top}{\sqrt{d_h}} \right) S \\
& Z^{v_e}_L = \text{LayerNorm} \left( Z^{v_e}_L + Z^{v_e} \right),
\end{aligned}
\end{equation}
where $d_h$ is hidden dimension for scaled attention, and $Z^{v_e}$ is the residual input for stabilization. The function $\text{LayerNorm}(\cdot)$ normalizes the output to improve training.

This adaptive fusion strategy enables each node to selectively integrate signals from various receptive field depths, improving the quality and robustness of local representations.

\subsection{Global Semantic Aggregation}
To supplement local representations with semantic-level global context, we introduce a cluster-aware global aggregation mechanism. A set of cluster centers, as global semantic prototypes, is dynamically updated. These centers are computed by applying a clustering algorithm (e.g., K-Means~\cite{hartigan1979algorithm} or spectral clustering~\cite{von2007tutorial}) to fused node representations \( Z^{v_e}_L \):

\begin{equation}
C_e = \Phi(Z^{v_e}_L) \in \mathbb{R}^{k \times d},
\end{equation}
where \( C_e \) denotes the set of \( k \) cluster centers derived from the current view \( e \), \( \Phi\) denotes a clustering algorithm and each center represents a global semantic anchor corresponding to a latent cluster.

To incorporate global semantics into node-level representations, a multi-head attention mechanism~\cite{vaswani2017attention} is used. In this process, node embeddings and global context are linearly projected and reshaped into multiple attention heads, forming the query, key, and value matrices:
\begin{equation} 
\begin{aligned}
& Q_G = \text{reshape}(Z^{v_e}_L W_G^q) \in \mathbb{R}^{n \times h \times d_h} \\
& K = V = \text{reshape}(C_e) \in \mathbb{R}^{k \times h \times d_h},
\end{aligned}
\end{equation}
where $W_G^q \in \mathbb{R}^{d \times d}$ is a learnable projection matrix for transforming node embeddings into queries, and $Z^{v_e}_L \in \mathbb{R}^{n \times d}$ represents the locally fused node embeddings from view $v_e$.

Scaled dot-product attention is then applied between each node and the cluster centers, enabling nodes to selectively attend to global semantic prototypes:

\begin{equation}
G = \text{softmax} \left( \frac{Q_G K^\top}{\sqrt{d_h}} \right) V,
\end{equation}
where $Q_G$, $K$, and $V$ are the query, key, and value matrices as defined in the previous step; $d_h$ denotes the dimensionality of each attention head.

Subsequently, a residual connection is formed between the global aggregated output and the locally fused representation, followed by layer normalization:

\begin{equation}
Z^{v_e}_G = \text{LayerNorm} \left( \beta \cdot G + (1 - \beta) \cdot Z^{v_e}_L \right),
\end{equation}
where $\beta \in [0, 1]$ is a coefficient controlling the trade-off between global and local signals. 

This global semantic aggregation mechanism strengthens node embeddings by incorporating high-level structural prototypes from the entire graph. Such enhancement is especially advantageous for nodes situated in weakly connected or isolated components, where local message passing is often insufficient for capturing informative global context.

\subsection{Representation Learning and Clustering Objective}
To obtain discriminative and semantically rich node representations, the representation learning loss $L_r$ consists of an instance-level contrastive term $L_{\text{infoNEC}}$ and a neighbor-supervised term $L_{\text{ns}}$.

The instance-level contrastive loss follows the InfoNCE principle, aiming to pull together representations of the same node across two augmented views while pushing apart representations of other node pairs. Given embeddings from the two augmented views \( Z^{v_e}_G \) and \( Z^{v_f}_G \), the similarity between two nodes is defined as:

\begin{equation}
\theta \left( i^{v_e}, j^{v_f} \right) = Z^{v_e}_{G_i} \cdot Z^{v_f}_{G_j},
\end{equation}
where $Z^{v_e}_{G_i}$ and $Z^{v_f}_{G_j}$ are the normalized embeddings of nodes $i$ and $j$ in views $v_e$ and $v_f$, respectively.

The instance-level InfoNCE loss is then given by:

\begin{equation}
L_{\text{infoNEC}} = \frac{1}{2N} \sum_{e=1}^{2} \sum_{i=1}^{N} L^{v_e}_i,
\end{equation}
{\small
\begin{equation}
L^{v_e}_i = -\log \frac{e^{\theta(i^{v_e}, i^{v_f})}}{e^{\theta(i^{v_e}, i^{v_f})} + \sum_{j \neq i} \left( e^{\theta(i^{v_e}, j^{v_e})} + e^{\theta(i^{v_e}, j^{v_f})} \right)}
\end{equation}
}
where \( L^{v_e}_i \) denotes the contrastive loss for node \( i \) in view \( v_e \), and \( N \) is the number of nodes.

To further incorporate structural signals, $L_{ns}$ is introduced. This term enforces similar embeddings for neighboring nodes based on the binarized normalized adjacency matrix \( \tilde{A} \):

{\small
\begin{align}
L_{\text{ns}} 
&= \frac{1}{N^2} \sum_{i=1}^{N} \sum_{j=1}^{N} \Bigg[
-\tilde{a}_{ij} \log\!\big(\theta(i^{v_e}, j^{v_f})\big) \notag \\
&\quad - (1 - \tilde{a}_{ij}) \log\!\big(1 - \theta(i^{v_e}, j^{v_f})\big)
\Bigg],
\end{align}
}
where \( \tilde{a}_{ij} \) indicates the presence of an edge between nodes \( i \) and \( j \).

The combined representation learning loss is defined as:

\begin{equation}
L_{\text{r}} = L_{\text{infoNEC}} + L_{\text{ns}}.
\end{equation}

Beyond improving embedding quality, a clustering alignment loss \( L_{\text{clu}} \) is adopted to enhance clustering performance, inspired by previous works~\cite{guo2017improved,wang2019attributed,xie2016unsupervised}. This term aligns the soft cluster assignments \( q \) with a sharpened target distribution \( p \), and is typically optimized using KL divergence.

Combining all components, the final training objective is:

\begin{equation}
L = L_{\text{r}} + \gamma L_{\text{clu}},
\end{equation}
where \( \gamma \) is a hyperparameter controlling the influence of the clustering objective.

\begin{table}[!th]
\centering

\resizebox{0.415\textwidth}{!}{ 
\begin{tabular}{lccccc}
\toprule
\textbf{Dataset}  & \textbf{\#Nodes} & \textbf{\#Edges} & \textbf{Edge Density} & \textbf{\#Features} & \textbf{\#Classes} \\ 
\midrule
UAT       & 1,190  & 13,599  & 0.0192  & 239  & 4   \\
Wiki      & 2,405  & 8,261   & 0.0029  & 4,973& 17  \\
Cora      & 2,708  & 5,278   & 0.0014  & 1,433& 7   \\
ACM       & 3,025  & 13,128  & 0.0029  & 1,870& 3   \\
Citeseer  & 3,327  & 4,552   & 0.0008  & 3,703& 6   \\
DBLP      & 4,057  & 3,528   & 0.0004  & 334  & 4   \\
AMAP      & 7,650  & 119,043 & 0.0041  & 745  & 8   \\
COCS      & 18,333 & 81,894  & 0.0005  & 6,805& 15  \\
\bottomrule
\end{tabular}
}
\caption{Dataset statistics of real-world datasets.}

\label{tab:dataset_stats}
\end{table}

\section{Experiments}
This section presents a comprehensive evaluation of the proposed model from multiple perspectives.
It begins by introducing benchmark datasets, evaluation metrics, and experimental setup.
The clustering results of RCLG are compared with state-of-the-art baselines, followed by analyses of RCLG variants, including ablation studies and hyperparameter sensitivity experiments.
Experiments were conducted on a workstation with an NVIDIA RTX 4090 GPU.

\begin{table*}[t]
\centering
\renewcommand{\arraystretch}{1.03}
\resizebox{0.95\textwidth}{!}{
\begin{tabular}{l|l|cccccccccc}
\hline
Dataset & Metrics & AGC & SDCN & FGC & VGAER & AGC-DRR & SCGC & MAGI & MCGC & WGCN & RCLG \\ \hline

\multirow{3}{*}{UAT}
& ACC & 48.72±0.40 & 43.85±2.28 & 45.65±2.70 & 48.02±0.80 & 41.38±3.73 & \textbf{59.74±0.82} & 53.72±0.38 & 54.14±0.38 & 58.04±0.91 & \underline{59.58±0.79} \\
& NMI & 23.74±1.17 & 13.59±2.19 & 23.20±1.87 & 24.44±0.88 & 10.58±2.79 & 29.32±1.06 & 21.74±0.85 & 27.20±0.26 & \underline{29.50±1.13} & \textbf{30.17±0.34} \\
& F1  & 44.57±0.93 & 40.72±2.37 & 40.17±5.00 & 49.13±0.62 & 37.22±5.29 & 57.33±1.47 & 52.04±0.46 & 55.61±0.20 & \underline{57.50±1.04} & \textbf{58.12±0.98} \\ \hline

\multirow{3}{*}{Wiki}
& ACC & 38.01±7.28 & 41.58±3.07 & 49.57±5.76 & 41.53±1.16 & 43.75±6.86 & 57.07±0.43 & 52.21±1.13 & 58.21±0.68 & \underline{58.44±0.84} & \textbf{59.13±0.70} \\
& NMI & 36.26±6.95 & 36.05±1.48 & 49.01±3.79 & 36.37±0.45 & 41.54±6.94 & 52.80±0.60 & 52.49±0.57 & 54.40±0.74 & \underline{55.06±0.97} & \textbf{55.16±0.60} \\
& F1  & 28.06±8.61 & 26.85±2.34 & 40.15±3.47 & 36.56±0.96 & 33.17±4.73 & 46.27±1.42 & 43.88±1.18 & \underline{48.71±0.56} & 48.66±0.82 & \textbf{49.98±0.77} \\ \hline

\multirow{3}{*}{Cora}
& ACC & 60.98±2.92 & 57.00±4.42 & 58.43±10.66 & 67.53±1.25 & 59.74±4.22 & 73.67±0.51 & \textbf{75.09±0.88} & 73.83±0.44 & 74.92±0.74 & \underline{75.06±0.38} \\
& NMI & 47.00±2.88 & 36.44±3.33 & 38.99±11.68 & 47.79±0.73 & 48.16±3.07 & 55.92±0.58 & 56.97±0.75 & 56.77±0.64 & \underline{58.52±0.96} & \textbf{59.18±0.72} \\
& F1  & 51.33±5.09 & 50.86±2.62 & 53.62±11.17 & 66.27±0.78 & 54.10±2.43 & 70.43±1.37 & \textbf{72.78±1.19} & 69.28±0.38 & 71.38±1.64 & \underline{72.61±0.76} \\ \hline  

\multirow{3}{*}{ACM}
& ACC & 83.50±0.00 & 74.50±2.13 & 81.55±1.76 & 51.04±0.49 & 87.50±10.03 & 89.91±0.32 & 89.58±0.37 & 89.34±0.31 & \underline{90.46±0.23} & \textbf{90.56±0.38} \\
& NMI & 55.17±0.06 & 44.23±1.65 & 51.94±3.84 & 18.56±0.59 & 63.11±17.05 & 66.75±0.64 & 65.58±0.64 & 66.10±0.93 & \underline{68.47±0.54} & \textbf{68.90±0.96} \\
& F1  & 83.60±0.00 & 74.26±2.39 & 81.01±1.92 & 41.31±0.64 & 87.45±10.21 & 89.87±0.34 & 89.65±0.35 & 89.30±0.24 & \underline{90.44±0.23} & \textbf{90.55±0.39} \\ \hline

\multirow{3}{*}{Citeseer}
& ACC & 68.43±0.05 & 62.03±1.83 & 61.20±3.36 & 47.46±0.99 & 67.71±1.35 & \underline{71.14±0.91} & 69.80±0.57 & 70.97±0.70 & \textbf{72.49±0.39} & 71.04±0.71 \\
& NMI & 42.61±0.03 & 35.55±2.14 & 35.34±3.98 & 25.16±0.80 & 42.67±1.43 & 44.78±1.08 & 43.96±0.52 & 44.87±0.30 & \textbf{46.76±0.59} & \underline{46.16±0.59} \\  
& F1  & 63.80±0.00 & 58.16±1.55 & 57.76±2.65 & 46.26±1.36 & 63.64±1.20 & 61.95±0.97 & \underline{65.03±0.70} & 61.89±0.54 & 63.34±0.49 & \textbf{65.92±0.57} \\ \hline

\multirow{3}{*}{DBLP}
& ACC & 62.38±3.51 & 56.31±0.94 & 73.65±0.98 & 44.54±1.88 & 61.43±2.34 & 68.23±0.58 & \underline{78.86±0.51} & 72.33±0.21 & 62.05±1.52 & \textbf{80.32±1.14} \\
& NMI & 32.15±3.20 & 23.64±1.78 & 42.51±1.69 & 11.57±1.26 & 32.77±2.29 & 39.86±0.57 & \underline{49.72±0.52} & 43.75±1.35 & 31.34±1.72 & \textbf{50.76±1.44} \\
& F1  & 62.12±3.53 & 55.74±1.20 & 73.31±0.97 & 43.35±1.81 & 59.13±2.07 & 68.02±0.60 & \underline{78.55±0.48} & 71.97±0.47 & 62.62±1.53 & \textbf{79.98±1.10} \\ \hline

\multirow{3}{*}{AMAP}
& ACC & 61.51±5.31 & 57.77±3.77 & 45.22±10.15 & 71.41±1.77 & 66.79±2.80 & 78.21±0.62 & \underline{79.01±0.47} & 77.81±0.65 & 78.76±0.62 & \textbf{83.66±2.17} \\
& NMI & 56.48±3.13 & 44.09±4.08 & 33.88±14.69 & 58.56±1.95 & 56.02±5.20 & 65.51±1.05 & \underline{70.23±0.67} & 65.23±0.31 & 68.65±0.76 & \textbf{75.03±2.18} \\
& F1  & 56.82±5.53 & 42.88±3.40 & 38.23±9.19 & 66.95±1.22 & 60.51±3.80 & 74.18±2.75 & 72.58±0.35 & \underline{75.39±0.35} & 72.76±0.59 & \textbf{83.20±2.32} \\ \hline

\multirow{3}{*}{COCS}
& ACC & 65.76±4.59 & 60.07±3.00 & 75.38±2.93 & 61.12±1.77 & 60.37±9.92 & 76.21±2.00 & 75.36±0.12 & 71.45±0.26 & \underline{76.58±1.27} & \textbf{81.85±1.89} \\ 
& NMI & 64.77±2.53 & 60.10±1.22 & 77.16±1.34 & 61.47±2.27 & 61.81±10.78 & 71.35±1.74 & 77.88±0.07 & 65.84±0.23 & \underline{78.44±1.26} & \textbf{82.90±0.56} \\ 
& F1  & 53.14±5.34 & 37.64±2.33 & \underline{75.71±3.90} & 57.60±2.68 & 43.10±7.60 & 69.08±3.26 & 74.08±0.08 & 55.02±0.32 & 70.71±0.69 & \textbf{76.14±1.70} \\ \hline

\end{tabular}
}
\caption{Clustering performance comparison on eight datasets (Mean ± Std over 10 runs; Bold = best; Underline = second best).}
\label{tab:clustering performance}
\end{table*}

\subsection{Benchmark Datasets and and Evaluation Metrics}
The evaluation is conducted on eight widely used real-world graph datasets with diverse structures and domains, covering air-traffic networks (UAT~\cite{mrabah2022rethinking}), word co-occurrence graphs (Wiki), citation networks (Cora~\cite{cui2020adaptive} and Citeseer~\cite{cui2020adaptive}), e-commerce co-purchase graphs (AMAP~\cite{liu2022deep}), and academic co-authorship networks (ACM~\cite{wang2019heterogeneous}, DBLP~\cite{wang2019heterogeneous}, and COCS~\cite{shchur2018pitfalls}).
Dataset statistics are summarized in Table~\ref{tab:dataset_stats}.  
Clustering performance is evaluated using three widely adopted metrics: Accuracy (ACC), Normalized Mutual Information (NMI), and macro F1-score (F1).

\subsection{Baselines and Experimental Setup}

Eight representative graph clustering methods are selected for comparison, including traditional structure modeling approaches and contrastive learning-based methods. Specifically, AGC~\cite{zhang2019attributed} employs adaptive graph convolution to model global structural information. SDCN~\cite{bo2020structural} integrates node attributes and structure in a unified clustering framework. FGC~\cite{kang2022fine} proposes a fine-grained strategy for attribute graph clustering. VGAER~\cite{qiu2022vgaer} utilizes variational inference to jointly model topology and attributes. AGC-DRR~\cite{gong2022attributed} designs a redundancy-reduced representation learning method. In addition, recent contrastive learning-based methods are considered, including SCGC~\cite{liu2023simple} which focuses on efficient neighbor contrast, MAGI~\cite{liu2024revisiting} which leverages modularity, MCGC~\cite{DBLP:journals/ipm/WuHQMS25} addressing motif-based contrastive clustering, and WGCN~\cite{wang2026representation} which adopts a representation-then-augmentation framework.

The model is trained with learning rate $\text{lr} = 5 \times 10^{-4}$ for 500 epochs. The noise perturbation coefficient is set to $\alpha = 0.01$, and the clustering update interval is $T = 5$. The global semantic fusion coefficient is $\beta = 0.2$, and the clustering alignment loss is weighted by $\gamma = 10$. The hidden representation dimensionality is 256, and the number of graph propagation steps is $l = 6$. For baselines, the best configuration is individually selected for each dataset, whereas our method uses a single fixed configuration across all datasets.

\subsection{Clustering Performance Comparisons}
As shown in Table~\ref{tab:clustering performance}, the proposed method exhibits particularly notable improvements on larger-scale datasets. Specifically, on the AMAP dataset, a relative accuracy improvement of approximately 5.88\% is achieved compared to the second-best approach, while on the COCS dataset, the improvement reaches 7.40\%. These results indicate that integrating local and global information becomes increasingly effective as the dataset scale increases. 
Moreover, the method is robust to varying sparsity. Table~\ref{tab:dataset_stats} reports that Wiki and ACM are relatively denser (0.0029), whereas DBLP (0.0004) and COCS (0.0005) are among the sparsest datasets; nevertheless, strong performance is consistently achieved, indicating reliable clustering under sparse or fragmented connectivity.

\subsection{Ablation Study}
To analyze the contribution of each key component and mechanism, ablation studies are conducted on different model variants. Specifically, w/o A denotes removing all attention mechanisms: attention is replaced with mean fusion in the local component, and equal weights are assigned to all cluster centers in the global component.  w/o L\&G denotes the variant that performs simple mean fusion of multi-scale representations without the global information integration module.  w/o G removes only the global information integration component, while RCLG represents the full model.
As shown in Table~\ref{tab:ablation}, on the Wiki dataset, the integration of local information improves clustering performance with a relative enhancement of 7.1\%. On the Cora dataset, incorporating global information yields a relative improvement of 1.7\%. These findings demonstrate that both local and global information integration modules substantially enhance the overall framework's effectiveness.

 In addition, Table~\ref{tab:ablation_loss} presents the effect of the two components constituting the representation learning loss $L_{\text{R}}$. The results indicate that both the instance-level contrastive loss $L_{\text{infoNEC}}$ and the neighbor-supervised contrastive loss $L_{\text{ns}}$ contribute to improving clustering performance. Among them, $L_{\text{infoNEC}}$ plays a more prominent role by ensuring fundamental discriminability between node embeddings, which is particularly beneficial for enhancing clustering quality.

\begin{table}[tbp]
\centering

\resizebox{0.45\textwidth}{!}{%
\begin{tabular}{l|l|cccccccc}
\hline
\textbf{Metrics} & \textbf{Methods} & \textbf{UAT} & \textbf{Wiki} & \textbf{Cora} & \textbf{ACM} & \textbf{Citeseer} & \textbf{DBLP} & \textbf{AMAP} & \textbf{COCS} \\
\hline
\multirow{4}{*}{ACC}
&  w/o A & 54.99 & 54.97 & 73.09 & 89.51 & 70.49 & 76.73 & 81.34 & 80.41 \\
&  w/o L\&G      & 55.06 & 54.67 & 73.26 & 89.55 & 69.80 & 77.27 & 81.34 & 80.91 \\
&  w/o G         & 58.74 & 58.54 & 73.80 & 90.01 & 70.38 & 79.70 & 82.88 & 81.09 \\
& RCLG               & \textbf{59.58} & \textbf{59.13} & \textbf{75.06} & \textbf{90.56} & \textbf{71.04} & \textbf{80.32} & \textbf{83.66} & \textbf{81.84} \\
\hline
\multirow{4}{*}{NMI}
&  w/o A & 25.90 & 48.92 & 57.74 & 66.19 & 44.75 & 45.85 & 73.72 & 79.38 \\
&  w/o L\&G      & 25.93 & 48.52 & 58.12 & 66.19 & 44.67 & 46.75 & 73.77 & 79.43 \\
&  w/o G         & 30.08 & 54.68 & 57.89 & 67.35 & 45.38 & 50.10 & 75.85 & 82.83 \\
& RCLG               & \textbf{30.17} & \textbf{55.16} & \textbf{59.18} & \textbf{68.90} & \textbf{46.16} & \textbf{50.76} & \textbf{75.03} & \textbf{82.90} \\
\hline
\multirow{4}{*}{F1}
&  w/o A & 54.40 & 46.22 & 70.76 & 89.49 & 65.54 & 76.47 & 82.31 & 77.73 \\
&  w/o L\&G      & 54.49 & 46.37 & 71.38 & 89.52 & 65.37 & 76.95 & 81.66 & 76.18 \\
&  w/o G         & 57.23 & 49.53 & 71.83 & 90.02 & 65.32 & 79.36 & 82.38 & \textbf{76.60} \\
& RCLG               & \textbf{58.12} & \textbf{49.98} & \textbf{72.61} & \textbf{90.55} & \textbf{65.92} & \textbf{79.98} & \textbf{83.20} & 76.14 \\
\hline
\end{tabular}%
}

\caption{ Comparison between RCLG and its variants in terms of ACC, NMI, and F1.}
\label{tab:ablation}
\vspace{-0.1cm}

\end{table}

\begin{table}[tbp]
\centering

\resizebox{0.466\textwidth}{!}{
\begin{tabular}{l|l|cccccccc}
\hline
Metrics & Methods & UAT & Wiki & Cora & ACM & Citeseer & DBLP & AMAP & COCS \\
\hline
\multirow{4}{*}{ACC}
& w/o $L_r$ & 50.08 & 54.15 & 24.85 & 69.49 & 25.62 & 57.88 & 66.16 & 69.69 \\
&  w/o $L_{InfoNEC}$ & 52.74 & 52.47 & 52.54 & 85.10 & 46.25 & 62.81 & 72.38 & 80.72 \\
&  w/o $L_{ns}$ & 58.47 & 55.92 & 74.16 & 89.65 & 69.20 & 68.78 & 82.37 & 78.98 \\
& RCLG & \textbf{59.58} & \textbf{59.13} & \textbf{75.06} & \textbf{90.56} & \textbf{71.04} & \textbf{80.32} & \textbf{83.66} & \textbf{81.85} \\
\hline

\multirow{4}{*}{NMI}
&  w/o $L_r$ & 25.09 & 47.54 & 5.40 & 29.45 & 2.40 & 26.59 & 56.66 & 73.95 \\
& w/o $L_{InfoNEC}$ & 25.73 & 50.09 & 33.43 & 57.71 & 23.22 & 31.62 & 60.66 & 77.90 \\
&  w/o $L_{ns}$ & 29.92 & 53.76 & 56.53 & 66.49 & 44.43 & 38.31 & 74.16 & 82.05 \\
& RCLG & \textbf{30.17} & \textbf{55.16} & \textbf{59.18} & \textbf{68.90} & \textbf{46.16} & \textbf{50.76} & \textbf{75.03} & \textbf{82.90} \\
\hline

\multirow{4}{*}{F1}
&  w/o $L_r$ & 48.49 & 41.96 & 22.53 & 69.26 & 22.90 & 58.36 & 62.76 & 65.24 \\
& w/o $L_{InfoNEC}$ & 50.47 & 44.43 & 51.56 & 85.06 & 42.67 & 62.77 & 68.35 & 69.01 \\
&  w/o $L_{ns}$ & 56.50 & 47.88 & 72.56 & 89.67 & 64.98 & 69.14 & 82.01 & 76.04 \\
& RCLG & \textbf{58.12} & \textbf{49.98} & \textbf{72.61} & \textbf{90.55} & \textbf{65.92} & \textbf{79.98} & \textbf{83.20} & \textbf{76.14} \\
\hline
\end{tabular}
}
\caption{Ablation Study on Instance-Level ($L_{\text{infoNEC}}$) and Neighbor-Supervised ($L_{\text{ns}}$) Contrastive Losses in the Full Representation Objective ($L_r$).}
\vspace{-0.4cm}
\label{tab:ablation_loss}
\end{table}

\begin{table}[tbp]
\centering
\resizebox{0.485\textwidth}{!}{%
\begin{tabular}{l|l|cccccccc}
\toprule
\textbf{Metrics} & \textbf{Model} & \textbf{UAT} & \textbf{Wiki} & \textbf{Cora} & \textbf{ACM} & \textbf{Citeseer} & \textbf{DBLP} & \textbf{AMAP} & \textbf{COCS} \\
\midrule
\multirow{2}{*}{\textbf{ACC}}     & SpecC & 58.12          & 59.10          & 74.74          & \textbf{91.22} & \textbf{71.06} & 80.16          & 83.27          & 80.21 \\
                                  & RCLG        & \textbf{59.58} & \textbf{59.13} & \textbf{75.06} & 90.56          & 71.04          & \textbf{80.32} & \textbf{83.66} & \textbf{81.85} \\
\midrule
\multirow{2}{*}{\textbf{NMI}}     & SpecC & 29.77          & \textbf{55.71} & 58.50          & \textbf{70.17} & \textbf{46.16} & \textbf{50.76} & \textbf{76.59} & 82.45 \\
                                  & RCLG        & \textbf{30.17} & 55.16          & \textbf{59.18} & 68.90          & \textbf{46.16} & \textbf{50.76} & 75.03          & \textbf{82.90} \\
\midrule
\multirow{2}{*}{\textbf{F1}}      & SpecC & 57.06          & 49.37          & 70.80          & \textbf{91.20} & \textbf{66.58} & 79.77          & 81.47          & 73.68 \\
                                  & RCLG        & \textbf{58.12} & \textbf{49.98} & \textbf{72.61} & 90.55          & 65.92          & \textbf{79.98} & \textbf{83.20} & \textbf{76.14} \\
\bottomrule
\end{tabular}
}

\caption{Performance comparison after replacing the clustering algorithm with Spectral Clustering (SpecC) method.}
\vspace{-0.1cm}

\label{tab:cluster_algo_comparison}
\end{table}

\subsection{Impact of Clustering Algorithm Choice}
To examine the effect of the clustering algorithm, we assess whether the Global component is sensitive to the choice of method. The Global module only requires reliable cluster centers to deliver global information via attention, and these centers are treated as learnable parameters updated jointly with the model, reducing dependence on random initialization and enhancing robustness.

To validate this, we replaced the default clustering algorithm with Spectral Clustering. As shown in Table~\ref{tab:cluster_algo_comparison}, the replacement has little impact on performance. Across datasets, Spectral Clustering and our original RCLG achieve comparable results, confirming that the model is not tied to a specific clustering algorithm and remains robust and effective.

\begin{figure}[!t]
    \centering
    \includegraphics[width=0.7\linewidth]{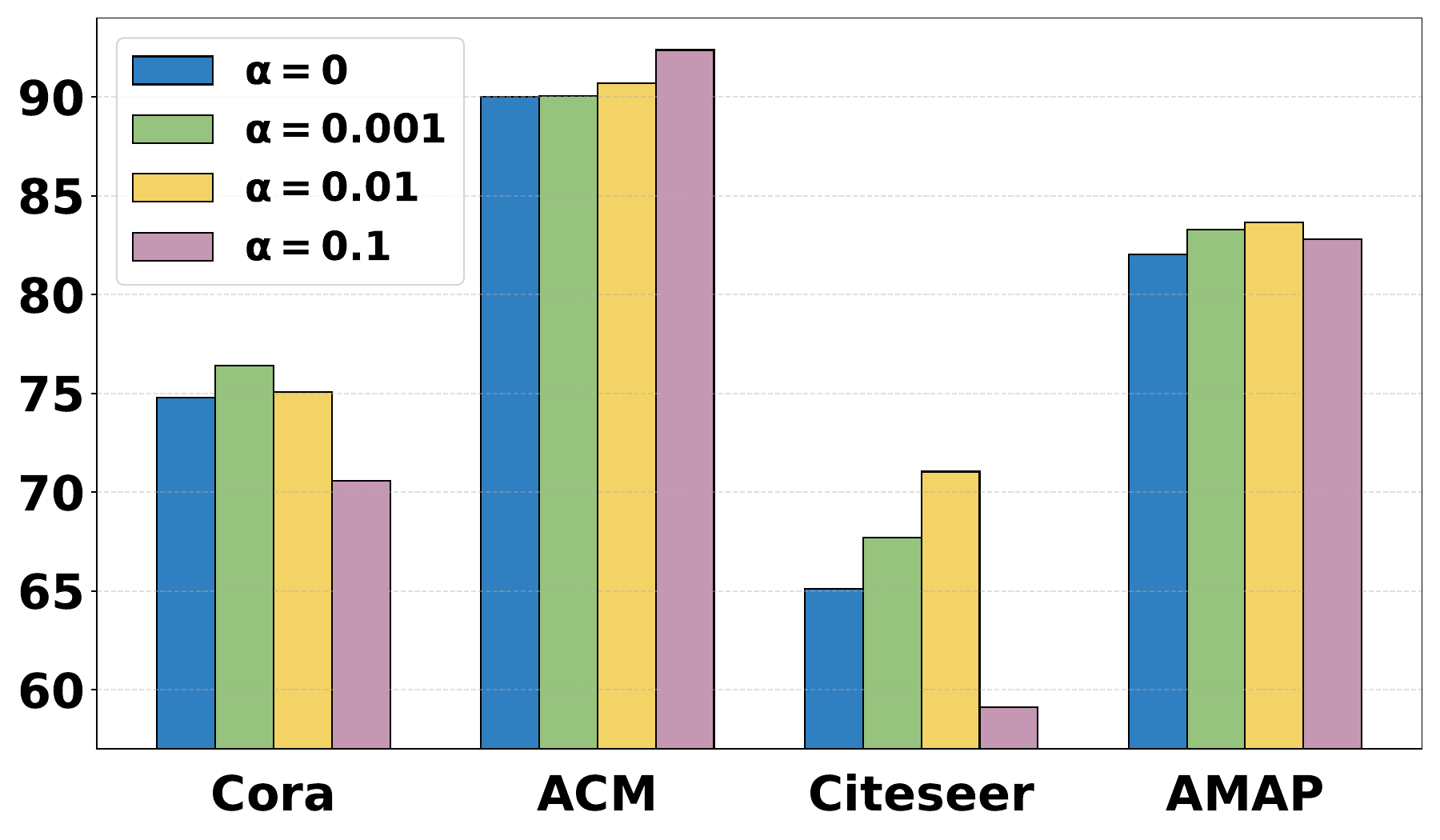}
    \caption{Sensitivity analysis of hyper-parameter noise coefficient $\alpha$ on clustering performance(ACC).}
    \label{fig:sensitivity_analysis}
    \vspace{-0.1cm}

\end{figure}

\begin{table}[!tbp]
\centering

\resizebox{0.48\textwidth}{!}{%
\begin{tabular}{l|cccccccccc}
\toprule
\textbf{Dataset} & \textbf{0} & \textbf{0.1} & \textbf{0.2} & \textbf{0.3} & \textbf{0.4} & \textbf{0.5} & \textbf{0.6} & \textbf{0.7} & \textbf{0.8} & \textbf{0.9} \\
\midrule
\textbf{UAT}      & 59.19          & 58.86          & 59.58          & \textbf{59.68} & 59.19          & 57.36          & 51.92          & 50.19          & 48.66          & 48.64          \\
\textbf{Wiki}     & 59.01          & 59.61          & 59.13          & 59.43          & 59.53          & \textbf{59.89} & 53.84          & 55.28          & 54.37          & 49.21          \\
\textbf{Cora}     & 72.48          & 73.35          & \textbf{75.06} & 73.50          & 71.86          & 52.85          & 33.29          & 26.22          & 26.61          & 25.57          \\
\textbf{ACM}      & 90.03          & 90.75          & 90.56          & 90.98          & \textbf{92.05} & 88.73          & 83.35          & 73.10          & 74.55          & 61.00          \\
\textbf{Citeseer} & 70.31          & 70.75          & \textbf{71.04} & 70.75          & 70.70          & 59.34          & 31.10          & 26.18          & 25.93          & 25.11          \\
\textbf{DBLP}     & 79.83          & 79.66          & \textbf{80.32} & 80.02          & 79.66          & 80.17          & 75.75          & 58.66          & 57.75          & 50.87          \\
\textbf{AMAP}     & 82.66          & 83.24          & \textbf{83.66} & 83.65          & 66.97          & 65.20          & 66.28          & 64.36          & 64.04          & 61.14          \\
\textbf{COCS}     & 81.09          & 81.33          & 81.85          & 81.43          & 84.50          & \textbf{87.09} & 84.16          & 80.65          & 75.44          & 74.06          \\
\bottomrule
\end{tabular}%
}

\caption{Sensitivity analysis of global fusion coefficient $\beta$ on clustering performance(ACC).}
\label{tab:sensitivity_beta}
        \vspace{-0.1cm}
\end{table}

\begin{table}[!tbp]
\centering
\resizebox{0.48\textwidth}{!}{%
\begin{tabular}{l|cccccccccc}
\toprule
\textbf{Dataset} & \textbf{1} & \textbf{2} & \textbf{3} & \textbf{4} & \textbf{5} & \textbf{6} & \textbf{7} & \textbf{8} & \textbf{9} & \textbf{10} \\
\midrule
\textbf{UAT}      & 57.51          & 58.77          & 58.71          & 59.63          & 59.10          & 59.58          & \textbf{59.97} & 58.71          & 59.49          & 58.97          \\
\textbf{Wiki}     & 59.83          & 59.27          & 57.96          & 59.00          & \textbf{60.36} & 59.13          & 58.62          & 59.20          & 58.45          & 58.68          \\
\textbf{Cora}     & 71.48          & 71.63          & 71.92          & 71.62          & 73.84          & \textbf{75.06} & 72.28          & 73.95          & 73.46          & 73.86          \\
\textbf{ACM}      & 90.23          & 90.77          & \textbf{91.01} & 90.81          & 90.75          & 90.56          & 90.14          & 90.27          & 90.46          & 90.07          \\
\textbf{Citeseer} & 70.43          & 70.34          & 70.89          & 71.50          & 70.95          & 71.04          & \textbf{71.38} & 71.01          & 71.01          & 70.50          \\
\textbf{DBLP}     & 80.09          & 79.66          & 80.26          & 80.32          & 79.48          & \textbf{80.35} & 79.88          & 79.32          & 78.26          & 79.73          \\
\textbf{AMAP}     & \textbf{84.62} & 83.04          & 84.62          & 83.13          & 82.49          & 83.66          & 83.91          & 82.39          & 83.26          & 82.01          \\
\textbf{COCS}     & 78.09          & 80.17          & 80.73          & 81.79          & 81.23          & 81.85          & 82.15          & 83.31          & 82.82          & \textbf{83.55} \\
\bottomrule
\end{tabular}%
}

\caption{Sensitivity of propagation steps $l$ on clustering performance(ACC).}
\label{tab:sensitivity_convs}
\vspace{-0.2cm}

\end{table}

\begin{figure}[t]

    \centering
    \includegraphics[width=0.45\textwidth]{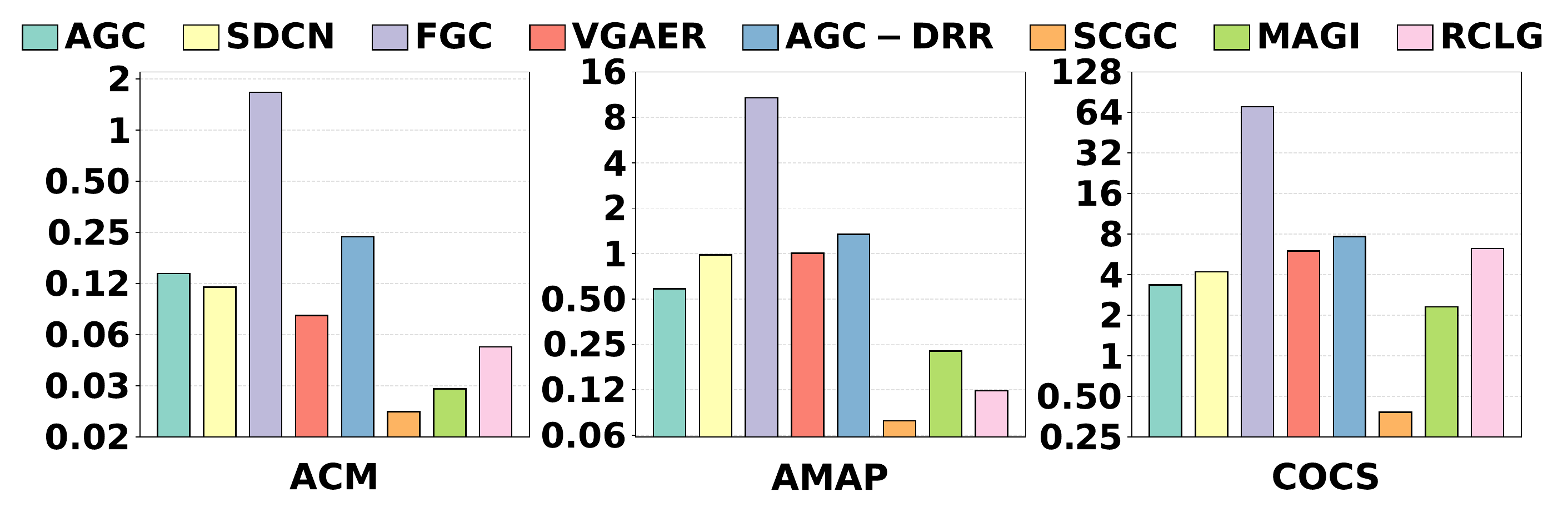}
    \caption{Per-epoch runtime comparison (s) of different methods across datasets.}
\vspace{-0.1cm}

    \label{fig:runtime_comparison}
\end{figure}
\subsection{Sensitivity Analysis}
To further assess model robustness, sensitivity analysis is performed on three hyperparameters: noise coefficient \( \alpha \), global fusion coefficient \( \beta \), and propagation layers \( l \).

As shown in Figure~\ref{fig:sensitivity_analysis}, setting the noise coefficient $\alpha$ to 0.01 yields stable and strong performance across multiple datasets, demonstrating the effectiveness of introducing mild noise to enhance representation robustness. Likewise, Table~\ref{tab:sensitivity_beta} shows that within the range $\beta \in [0.2, 0.4]$, each dataset tends to achieve its optimal performance, indicating that the model remains robust and broadly applicable when balancing global and local information in this interval. When either \( \alpha \) or \( \beta \) is set excessively high, clustering performance degrades on several datasets, suggesting that overly strong perturbations or global dependencies may overwhelm local learning. This confirms the necessity of maintaining a balanced integration between local and global information.

Furthermore, as summarized in Table~\ref{tab:sensitivity_convs}, datasets with larger diameters generally reach best performance at slightly deeper propagation steps, reflecting the need for sufficient receptive field expansion to capture long-range dependencies. Considering both accuracy and efficiency, we fix the propagation depth to \( l = 6 \) in all subsequent experiments.

\
\subsection{Computational Efficiency}
To further assess the practicality of our method, we compared the per-epoch runtime of RCLG with several state-of-the-art baselines on three representative datasets, namely COCS (18,333 nodes), AMAP (7,650 nodes), and ACM (3,025 nodes). As summarized in Figure~\ref{fig:runtime_comparison}, the runtime of RCLG is of the same order of magnitude as other methods, indicating that the improved clustering performance does not incur excessive computational overhead. This suggests that our framework achieves competitive efficiency while maintaining superior clustering quality.

\section{Conclusion}
This paper presents a robust contrastive graph clustering framework that leverages both local and global structural signals within a dual-view architecture to address key limitations. A multi-level attention fusion aggregates neighborhood information from multiple convolutional layers, allowing each node to selectively absorb informative cues across depths and reducing noise from rigid message passing. Semantic prototypes from evolving cluster centers further provide global guidance, enabling nodes—especially in sparse or disconnected regions—to incorporate context beyond immediate neighbors, stabilizing long-range semantics and strengthening cross-community discrimination. Together, these mechanisms improve representation quality and promote more compact and separable clusters. The framework achieves strong performance across diverse benchmarks without dataset-specific hyperparameter tuning.

\section*{Acknowledgements}
This work was supported by the National Natural Science Foundation of China (No. 61976001, 62502340), the Natural Science Foundation of Anhui Province (No. 2408085MF152, 2508085MF176) and the Natural Science Foundation of Tianjin (No. 24JCQNJC01560).

\bibliographystyle{named}
\bibliography{ijcai26}

@String{Computing = "Computing" }

@String{Springer = "Springer-Verlag" }

@article{liu2023simple,
  title={Simple contrastive graph clustering},
  author={Liu, Yue and Yang, Xihong and Zhou, Sihang and Liu, Xinwang and Wang, Siwei and Liang, Ke and Tu, Wenxuan and Li, Liang},
  journal={IEEE Transactions on Neural Networks and Learning Systems},
  year={2023},
  publisher={IEEE}
}

@article{hartigan1979algorithm,
  title={Algorithm AS 136: A k-means clustering algorithm},
  author={Hartigan, John A and Wong, Manchek A},
  journal={Journal of the Royal Statistical Society. series c (applied statistics)},
  volume={28},
  number={1},
  pages={100--108},
  year={1979},
  publisher={JSTOR}
}

@article{von2007tutorial,
  title={A tutorial on spectral clustering},
  author={Von Luxburg, Ulrike},
  journal={Statistics and Computing},
  volume={17},
  pages={395--416},
  year={2007},
  publisher={Springer}
}

@article{wang2019attributed,
  title={Attributed graph clustering: A deep attentional embedding approach},
  author={Wang, Chun and Pan, Shirui and Hu, Ruiqi and Long, Guodong and Jiang, Jing and Zhang, Chengqi},
  journal={arXiv preprint arXiv:1906.06532},
  year={2019}
}

@inproceedings{xie2016unsupervised,
  title={Unsupervised deep embedding for clustering analysis},
  author={Xie, Junyuan and Girshick, Ross and Farhadi, Ali},
  booktitle={International Conference on Machine Learning},
  pages={478--487},
  year={2016},
  organization={PMLR}
}

@inproceedings{guo2017improved,
  title={Improved deep embedded clustering with local structure preservation},
  author={Guo, Xifeng and Gao, Long and Liu, Xinwang and Yin, Jianping},
  booktitle={IJCAI},
  volume={17},
  pages={1753--1759},
  year={2017}
}

@inproceedings{liu2024revisiting,
  title={Revisiting modularity maximization for graph clustering: A contrastive learning perspective},
  author={Liu, Yunfei and Li, Jintang and Chen, Yuehe and Wu, Ruofan and Wang, Ericbk and Zhou, Jing and Tian, Sheng and Shen, Shuheng and Fu, Xing and Meng, Changhua and others},
  booktitle={Proceedings of the 30th ACM SIGKDD International Conference on Knowledge Discovery \& Data Mining},
  pages={1968--1979},
  year={2024}
}

@article{zhang2019attributed,
  title={Attributed graph clustering via adaptive graph convolution},
  author={Zhang, Xiaotong and Liu, Han and Li, Qimai and Wu, Xiao-Ming},
  journal={arXiv preprint arXiv:1906.01210},
  year={2019}
}

@inproceedings{bo2020structural,
  title={Structural deep clustering network},
  author={Bo, Deyu and Wang, Xiao and Shi, Chuan and Zhu, Meiqi and Lu, Emiao and Cui, Peng},
  booktitle={Proceedings of the Web Conference 2020},
  pages={1400--1410},
  year={2020}
}

@inproceedings{kang2022fine,
  title={Fine-grained attributed graph clustering},
  author={Kang, Zhao and Liu, Zhanyu and Pan, Shirui and Tian, Ling},
  booktitle={Proceedings of the 2022 SIAM International Conference on Data Mining (SDM)},
  pages={370--378},
  year={2022},
  organization={SIAM}
}

@article{qiu2022vgaer,
  title={VGAER: graph neural network reconstruction based community detection},
  author={Qiu, Chenyang and Huang, Zhaoci and Xu, Wenzhe and Li, Huijia},
  journal={arXiv preprint arXiv:2201.04066},
  year={2022}
}

@inproceedings{gong2022attributed,
  title={Attributed Graph Clustering with Dual Redundancy Reduction.},
  author={Gong, Lei and Zhou, Sihang and Tu, Wenxuan and Liu, Xinwang},
  booktitle={IJCAI},
  pages={3015--3021},
  year={2022}
}

@inproceedings{wang2019heterogeneous,
  title={Heterogeneous graph attention network},
  author={Wang, Xiao and Ji, Houye and Shi, Chuan and Wang, Bai and Ye, Yanfang and Cui, Peng and Yu, Philip S},
  booktitle={The World Wide Web Conference},
  pages={2022--2032},
  year={2019}
}

@article{shchur2018pitfalls,
  title={Pitfalls of graph neural network evaluation},
  author={Shchur, Oleksandr and Mumme, Maximilian and Bojchevski, Aleksandar and G{\"u}nnemann, Stephan},
  journal={arXiv preprint arXiv:1811.05868},
  year={2018}
}

@inproceedings{qiu2018deepinf,
  title={DeepInf: Modeling influence locality in large social networks},
  author={Qiu, Jiezhong and Tang, Jian and Ma, Hao and Dong, Yuxiao and Wang, Kuansan and Tang, Jie},
  booktitle={Proceedings of the 24th ACM SIGKDD International Conference on Knowledge Discovery \& Data Mining},
  pages={2110--2119},
  year={2018}
}

@article{BELLOMARINI2022101528,
title = {Vadalog: A modern architecture for automated reasoning with large knowledge graphs},
journal = {Information Systems},
volume = {105},
pages = {101528},
year = {2022},
issn = {0306-4379},
author = {Luigi Bellomarini and Davide Benedetto and Georg Gottlob and Emanuel Sallinger},
}

@inproceedings{fan2019graph,
  title={Graph neural networks for social recommendation},
  author={Fan, Wenqi and Ma, Yao and Li, Qing and He, Yuan and Zhao, Eric and Tang, Jiliang and Yin, Dawei},
  booktitle={The World Wide Web Conference},
  pages={417--426},
  year={2019}
}

@inproceedings{tu2021deep,
  title={Deep fusion clustering network},
  author={Tu, Wenxuan and Zhou, Sihang and Liu, Xinwang and Guo, Xifeng and Cai, Zhiping and Zhu, En and Cheng, Jieren},
  booktitle={Proceedings of the AAAI Conference on Artificial Intelligence},
  volume={35},
  number={11},
  pages={9978--9987},
  year={2021}
}

@article{peng2024multi,
  title={Multi-view graph imputation network},
  author={Peng, Xin and Cheng, Jieren and Tang, Xiangyan and Zhang, Bin and Tu, Wenxuan},
  journal={Information Fusion},
  volume={102},
  pages={102024},
  year={2024},
  publisher={Elsevier}
}

@article{kipf2016semi,
  title={Semi-supervised classification with graph convolutional networks},
  author={Kipf, Thomas N and Welling, Max},
  journal={arXiv preprint arXiv:1609.02907},
  year={2016}
}

@inproceedings{S3GC,
  author       = {Devvrit and
                  Aditya Sinha and
                  Inderjit S. Dhillon and
                  Prateek Jain},
  editor       = {Sanmi Koyejo and
                  S. Mohamed and
                  A. Agarwal and
                  Danielle Belgrave and
                  K. Cho and
                  A. Oh},
  title        = {{S3GC:} Scalable Self-Supervised Graph Clustering},
  booktitle    = {Advances in Neural Information Processing Systems 35: Annual Conference
                  on Neural Information Processing Systems 2022, NeurIPS 2022, New Orleans,
                  LA, USA, November 28 - December 9, 2022},
  year         = {2022},

}

@inproceedings{DBLP:conf/mm/MaZ23,
  author       = {Yixuan Ma and
                  Kun Zhan},
  editor       = {Abdulmotaleb El{-}Saddik and
                  Tao Mei and
                  Rita Cucchiara and
                  Marco Bertini and
                  Diana Patricia Tobon Vallejo and
                  Pradeep K. Atrey and
                  M. Shamim Hossain},
  title        = {Self-Contrastive Graph Diffusion Network},
  booktitle    = {Proceedings of the 31st {ACM} International Conference on Multimedia,
                  {MM} 2023, Ottawa, ON, Canada, 29 October 2023- 3 November 2023},
  pages        = {3857--3865},
  publisher    = {{ACM}},
  year         = {2023},
}

@inproceedings{CCGC,
  title={Cluster-guided Contrastive Graph Clustering Network},
  author={Yang, Xihong and Liu, Yue and Zhou, Sihang and Wang, Siwei and Tu, Wenxuan and Zheng, Qun and Liu, Xinwang and Fang, Liming and Zhu, En},
  booktitle={Proceedings of the AAAI Conference on Artificial Intelligence},
  volume={37},
  number={9},
  pages={10834--10842},
  year={2023}
}

@inproceedings{peng2021attention,
  title={Attention-driven graph clustering network},
  author={Peng, Zhihao and Liu, Hui and Jia, Yuheng and Hou, Junhui},
  booktitle={Proceedings of the 29th ACM International Conference on Multimedia},
  pages={935--943},
  year={2021}
}

@inproceedings{liu2020towards,
  title={Towards deeper graph neural networks},
  author={Liu, Meng and Gao, Hongyang and Ji, Shuiwang},
  booktitle={Proceedings of the 26th ACM SIGKDD International Conference on Knowledge Discovery \& Data Mining},
  pages={338--348},
  year={2020}
}

@inproceedings{xu2018representation,
  title={Representation learning on graphs with jumping knowledge networks},
  author={Xu, Keyulu and Li, Chengtao and Tian, Yonglong and Sonobe, Tomohiro and Kawarabayashi, Ken-ichi and Jegelka, Stefanie},
  booktitle={International Conference on Machine Learning},
  pages={5453--5462},
  year={2018},
  organization={PMLR}
}

@article{li2024multi,
  title={Multi-scale graph clustering network},
  author={Li, Xiulai and Wu, Wei and Zhang, Bin and Peng, Xin},
  journal={Information Sciences},
  volume={678},
  pages={121023},
  year={2024},
  publisher={Elsevier}
}

@article{vaswani2017attention,
  title={Attention is all you need},
  author={Vaswani, Ashish and Shazeer, Noam and Parmar, Niki and Uszkoreit, Jakob and Jones, Llion and Gomez, Aidan N and Kaiser, {\L}ukasz and Polosukhin, Illia},
  journal={Advances in Neural Information Processing Systems},
  volume={30},
  year={2017}
}

@article{xu2021infogcl,
  title={Infogcl: Information-aware graph contrastive learning},
  author={Xu, Dongkuan and Cheng, Wei and Luo, Dongsheng and Chen, Haifeng and Zhang, Xiang},
  journal={Advances in Neural Information Processing Systems},
  volume={34},
  pages={30414--30425},
  year={2021}
}

@inproceedings{xia2022simgrace,
  title={Simgrace: A simple framework for graph contrastive learning without data augmentation},
  author={Xia, Jun and Wu, Lirong and Chen, Jintao and Hu, Bozhen and Li, Stan Z},
  booktitle={Proceedings of the ACM Web Conference 2022},
  pages={1070--1079},
  year={2022}
}

@article{zhang2022enhancing,
  title={Enhancing sequential recommendation with graph contrastive learning},
  author={Zhang, Yixin and Liu, Yong and Xu, Yonghui and Xiong, Hao and Lei, Chenyi and He, Wei and Cui, Lizhen and Miao, Chunyan},
  journal={arXiv preprint arXiv:2205.14837},
  year={2022}
}

@inproceedings{he2022block,
  title={Block modeling-guided graph convolutional neural networks},
  author={He, Dongxiao and Liang, Chundong and Liu, Huixin and Wen, Mingxiang and Jiao, Pengfei and Feng, Zhiyong},
  booktitle={Proceedings of the AAAI Conference on Artificial Intelligence},
  volume={36},
  number={4},
  pages={4022--4029},
  year={2022}
}

@inproceedings{wang2022powerful,
  title={Powerful graph convolutional networks with adaptive propagation mechanism for homophily and heterophily},
  author={Wang, Tao and Jin, Di and Wang, Rui and He, Dongxiao and Huang, Yuxiao},
  booktitle={Proceedings of the AAAI Conference on Artificial Intelligence},
  volume={36},
  number={4},
  pages={4210--4218},
  year={2022}
}

@article{mrabah2022rethinking,
  title={Rethinking graph auto-encoder models for attributed graph clustering},
  author={Mrabah, Nairouz and Bouguessa, Mohamed and Touati, Mohamed Fawzi and Ksantini, Riadh},
  journal={IEEE Transactions on Knowledge and Data Engineering},
  volume={35},
  number={9},
  pages={9037--9053},
  year={2022},
  publisher={IEEE}
}

@inproceedings{cui2020adaptive,
  title={Adaptive graph encoder for attributed graph embedding},
  author={Cui, Ganqu and Zhou, Jie and Yang, Cheng and Liu, Zhiyuan},
  booktitle={Proceedings of the 26th ACM SIGKDD International Conference on Knowledge Discovery \& Data Mining},
  pages={976--985},
  year={2020}
}

@inproceedings{liu2022deep,
  title={Deep graph clustering via dual correlation reduction},
  author={Liu, Yue and Tu, Wenxuan and Zhou, Sihang and Liu, Xinwang and Song, Linxuan and Yang, Xihong and Zhu, En},
  booktitle={Proceedings of the AAAI Conference on Artificial Intelligence},
  volume={36},
  number={7},
  pages={7603--7611},
  year={2022}
}

@inproceedings{KunDGAC2025,
  author       = {Kun Xie and
                  Renchi Yang and
                  Sibo Wang},
  title        = {Diffusion-based Graph-agnostic Clustering},
  booktitle    = {Proceedings of the {ACM} on Web Conference 2025, {WWW} 2025, Sydney,
                  NSW, Australia, 28 April 2025- 2 May 2025},
  pages        = {1353--1364},
  publisher    = {{ACM}},
  year         = {2025},
}

@inproceedings{DBLP:conf/www/ZhaoLZWZL25,
  author       = {Peiyao Zhao and
                  Xin Li and
                  Zeyu Zhang and
                  Mingzhong Wang and
                  Xueying Zhu and
                  Lejian Liao},
  title        = {Robust Deep Signed Graph Clustering via Weak Balance Theory},
  booktitle    = {Proceedings of the {ACM} on Web Conference 2025, {WWW} 2025, Sydney,
                  NSW, Australia, 28 April 2025- 2 May 2025},
  pages        = {3819--3830},
  publisher    = {{ACM}},
  year         = {2025},
}

@inproceedings{DBLP:conf/www/WuLCLCNJJS025,
  author       = {Xinyi Wu and
                  Donald Loveland and
                  Runjin Chen and
                  Yozen Liu and
                  Xin Chen and
                  Leonardo Neves and
                  Ali Jadbabaie and
                  Mingxuan Ju and
                  Neil Shah and
                  Tong Zhao},
  title        = {GraphHash: Graph Clustering Enables Parameter Efficiency in Recommender
                  Systems},
  booktitle    = {Proceedings of the {ACM} on Web Conference 2025, {WWW} 2025, Sydney,
                  NSW, Australia, 28 April 2025- 2 May 2025},
  pages        = {357--369},
  publisher    = {{ACM}},
  year         = {2025},
}

@article{DBLP:journals/ipm/WuHQMS25,
  author       = {Xunlian Wu and
                  Jingqi Hu and
                  Yining Quan and
                  Qiguang Miao and
                  Peng Gang Sun},
  title        = {Motif-based Contrastive Graph Clustering with clustering-oriented
                  prompt},
  journal      = {Information Processing \& Management},
  volume       = {62},
  number       = {5},
  pages        = {104208},
  year         = {2025},

}

@inproceedings{peng2025sola,
  title={SOLA-GCL: Subgraph-Oriented Learnable Augmentation Method for Graph Contrastive Learning},
  author={Peng, Tianhao and Li, Xuhong and Yuan, Haitao and Li, Yuchen and Xiong, Haoyi},
  booktitle={Proceedings of the AAAI Conference on Artificial Intelligence},
  volume={39},
  number={19},
  pages={19875--19883},
  year={2025}
}

@inproceedings{DBLP:conf/kdd/ZhengZLCCZW25,
  author       = {Bowen Zheng and
                  Junjie Zhang and
                  Hongyu Lu and
                  Yu Chen and
                  Ming Chen and
                  Wayne Xin Zhao and
                  Ji{-}Rong Wen},
  title        = {Enhancing Graph Contrastive Learning with Reliable and Informative
                  Augmentation for Recommendation},
  booktitle    = {Proceedings of the 31st ACM SIGKDD International Conference on Knowledge Discovery
                  \& Data Mining},

  pages        = {2101--2112},
  publisher    = {{ACM}},
  year         = {2025},
}

@article{wang2026representation,
  author       = {Youqing Wang and
                  Tianxiang Zhao and
                  Mingliang Cui and
                  Junbin Gao and
                  Li Liang and
                  Jipeng Guo},
  title        = {Representation Then Augmentation: Wide Graph Clustering Network with
                  Multi-Order Filter Fusion and Double-Level Contrastive Learning},
  journal={IEEE/CAA Journal of Automatica Sinica},
  volume       = {13},
  number       = {2},
  pages        = {421--435},
  year         = {2026},
}

\end{document}


\maketitle
\appendix
\section{Appendix}
This appendix provides additional analyses and qualitative evidence to complement the main results. The following supplementary experiments are included to better understand the behavior and design choices of RCLG. First, augmentation strategies are compared to verify the effectiveness of the proposed noise-based augmentation against common structural perturbations. Second, the influence of contrastive pair construction in the neighbor-supervised loss $L_{\mathrm{ns}}$ is examined by testing different pair graphs. Third, visualization results are reported to offer an intuitive understanding of how local--global fusion progressively improves clustering quality.

\subsection{Analysis of Augmentation Strategies}

To assess the effectiveness of our noise-based augmentation, we compare RCLG with several widely used structural augmentation strategies. Many graph contrastive learning methods rely on edge manipulations or diffusion to create diverse views, but these approaches often introduce extra hyperparameters and non-trivial operations, and may also distort the original graph topology. For fairness, we benchmark RCLG against three representative augmentations: randomly dropping 10\% edges (“Drop”), randomly adding 10\% edges (“Add”), and graph diffusion (“Diffusion”) with a 0.20 teleportation rate.

As shown in Table~\ref{tab:augmentation_comparison}, RCLG achieves the best overall performance on most datasets and metrics, with particularly large gains on Citeseer and AMAP. This suggests that feature-space perturbations can provide sufficiently strong view diversity without explicitly modifying edges, and can be more robust when the observed graph structure is sparse or noisy. The results are consistent with observations in SCGC, which also highlights the effectiveness of feature-level augmentation for graph contrastive learning. In addition, the comparison reveals practical limitations of diffusion-based augmentation: applying diffusion on the largest dataset triggers Out-Of-Memory (OOM) errors, whereas the proposed noise-based strategy remains lightweight and scalable. Interestingly, while edge Drop/Add can occasionally achieve competitive results (e.g., UAT and COCS for some metrics), such gains are not consistent across datasets and may be sensitive to the chosen perturbation ratio. Overall, these findings indicate that a simple feature noise augmentation offers a strong accuracy--efficiency trade-off, serving as an effective and stable alternative to more complex structural transformations.

\begin{table}[!tbp]
\centering
\resizebox{0.46\textwidth}{!}{%
\begin{tabular}{l|l|cccc}
\toprule
\textbf{Dataset} & \textbf{Metrics} & \textbf{Drop } & \textbf{Add } & \textbf{Diffusion} & \textbf{RCLG} \\
\midrule
\multirow{3}{*}{\textbf{UAT}}      & ACC & 59.16 & \textbf{59.90} & 55.30 & 59.58 \\
                                   & NMI & 30.48 & \textbf{30.97} & 27.49 & 30.17 \\
                                   & F1  & 58.09 & \textbf{58.43} & 53.86 & 58.12 \\
\midrule
\multirow{3}{*}{\textbf{Wiki}}     & ACC & 58.36 & 57.84 & 58.16 & \textbf{59.13} \\
                                   & NMI & 55.16 & 54.37 & 54.54 & \textbf{55.16} \\
                                   & F1  & 49.68 & 49.39 & 48.43 & \textbf{49.98} \\
\midrule
\multirow{3}{*}{\textbf{Cora}}     & ACC & 73.20 & 74.07 & 74.99 & \textbf{75.06} \\
                                   & NMI & 55.42 & 56.29 & 57.68 & \textbf{59.18} \\
                                   & F1  & 71.52 & 72.36 & \textbf{73.38} & 72.61 \\
\midrule
\multirow{3}{*}{\textbf{ACM}}      & ACC & 90.50 & 89.75 & 90.53 & \textbf{90.56} \\
                                   & NMI & 68.22 & 66.33 & 68.72 & \textbf{68.90} \\
                                   & F1  & 90.51 & 89.75 & 90.54 & \textbf{90.55} \\
\midrule
\multirow{3}{*}{\textbf{Citeseer}} & ACC & 64.64 & 64.03 & 65.07 & \textbf{71.04} \\
                                   & NMI & 40.33 & 39.05 & 40.37 & \textbf{46.16} \\
                                   & F1  & 61.53 & 60.08 & 61.24 & \textbf{65.92} \\
\midrule
\multirow{3}{*}{\textbf{DBLP}}     & ACC & 79.68 & 78.20 & 79.02 & \textbf{80.32} \\
                                   & NMI & 49.59 & 47.81 & 49.03 & \textbf{50.76} \\
                                   & F1  & 79.34 & 77.69 & 78.56 & \textbf{79.98} \\
\midrule
\multirow{3}{*}{\textbf{AMAP}}     & ACC & 81.65 & 79.99 & 81.22 & \textbf{83.66} \\
                                   & NMI & 73.09 & 71.80 & 71.87 & \textbf{75.03} \\
                                   & F1  & 81.59 & 79.11 & 79.25 & \textbf{83.20} \\
\midrule
\multirow{3}{*}{\textbf{COCS}}     & ACC & \textbf{85.04}
 & 83.86
 & \multicolumn{1}{c}{OOM} & 81.85\\
                                   & NMI & \textbf{83.91}
 & 83.22
 & \multicolumn{1}{c}{OOM} & 82.90 \\
                                   & F1  & 77.73
 & \textbf{79.3}
 & \multicolumn{1}{c}{OOM} & 76.14 \\
\bottomrule
\end{tabular}%
}
\caption{Performance comparison with other graph augmentation strategies.}
\label{tab:augmentation_comparison}

\end{table}

\begin{figure*}[t]
    \centering
    \subfigure[DBLP]{
        \includegraphics[width=0.47\textwidth]{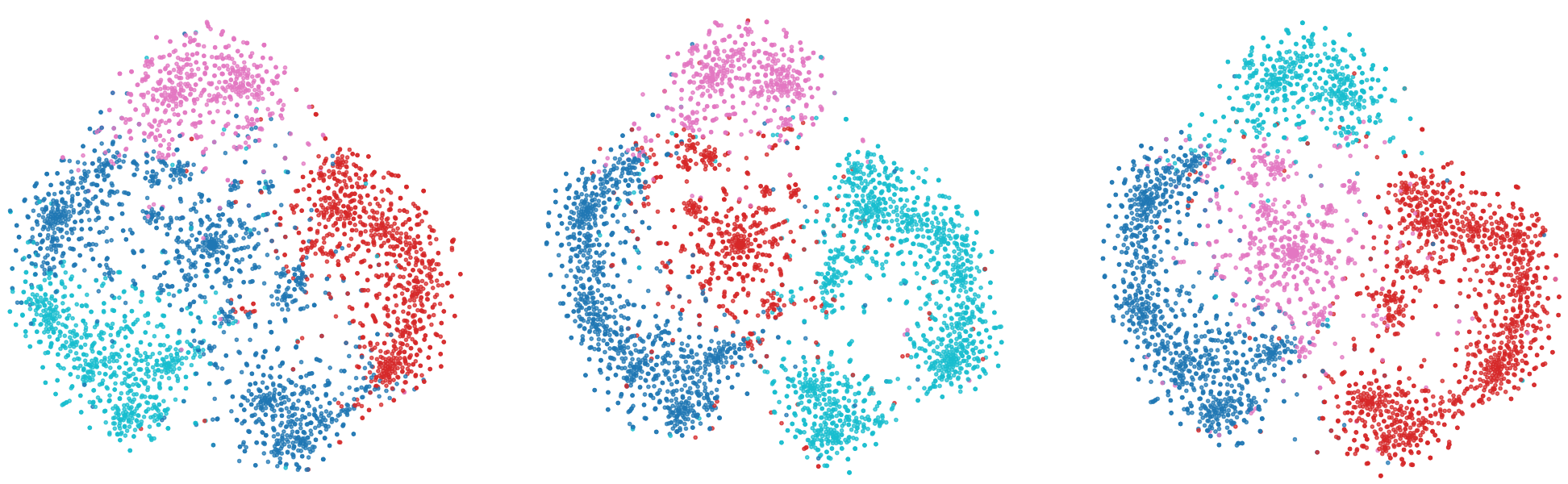}
    } \hfill
    \subfigure[AMAP]{
        \includegraphics[width=0.47\textwidth]{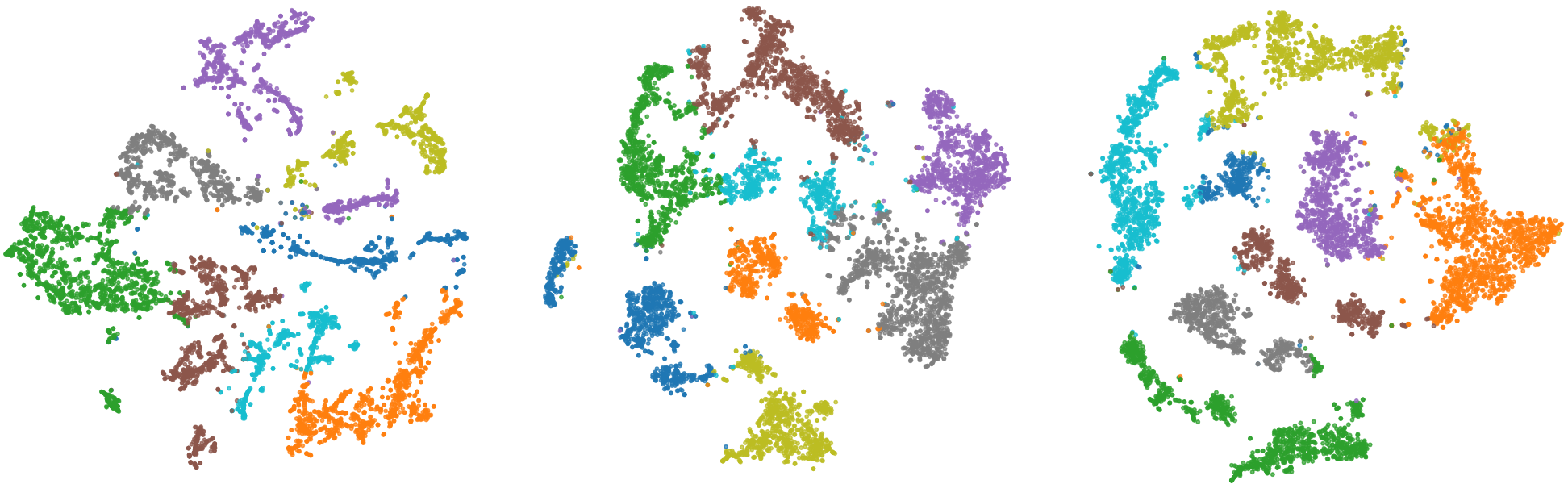}
    }
    \caption{Clustering visualization of node embeddings under different fusion strategies on DBLP and AMAP. From left to right: Single-Layer, Local Fusion, and Local + Global Fusion.}
    \label{fig:cluster_visualization}
\end{figure*}

\subsection{Analysis of Contrastive Pair Generation Strategies}
To further examine the impact of positive/negative pair construction in contrastive graph clustering, additional experiments are conducted by varying the pair generation strategy used in the neighbor-supervised loss $L_{\mathrm{ns}}$. In RCLG, $L_{\mathrm{ns}}$ uses a pair graph to define structural neighbors as positives, which is instantiated by the observed adjacency by default. To study how the choice of pair graph affects performance, three strategies are compared: (1) \emph{Random}, where node pairs are randomly linked to form supervision edges; (2) \emph{KNN}, where a $k$-Nearest Neighbors graph (top-10) is constructed based on learned node embeddings to connect semantically similar nodes; and (3) \emph{RCLG}, which uses the original adjacency matrix for structural sampling.

As shown in Table~\ref{tab:pair_generation}, replacing adjacency-based sampling with KNN can yield higher performance on certain datasets, indicating that similarity-based neighborhoods may provide stronger positive pairs when graph connectivity is weak. Nevertheless, the adjacency-based strategy remains attractive due to its simplicity and efficiency, avoiding the additional computation and design choices required to construct a KNN graph, while incurring only minor performance loss overall. In contrast, random pair selection consistently leads to degraded results, highlighting that informative pair selection is critical for effective contrastive learning.

\begin{table}[t]
\centering
\small
\setlength{\tabcolsep}{4pt}
\renewcommand{\arraystretch}{1.05}
\resizebox{\linewidth}{!}{
\begin{tabular}{c|c|ccccccc}
\hline
Metrics & Strategies & UAT & Wiki & Cora & ACM & Citeseer & DBLP & AMAP \\
\hline
\multirow{3}{*}{ACC} 
& Random & 58.72 & 58.63 & 73.14 & 90.26 & 70.06 & 79.19 & 81.93 \\
& KNN    & 59.23 & 58.76 & 74.09 & \textbf{90.89} & 70.78 & 79.72 & \textbf{83.89} \\
& RCLG   & \textbf{59.58} & \textbf{59.13} & \textbf{75.06} & 90.56 & \textbf{71.04} & \textbf{80.32} & 83.66 \\
\hline
\multirow{3}{*}{NMI} 
& Random & 29.87 & 54.47 & 57.81 & 68.05 & 45.17 & 49.81 & 74.33 \\
& KNN    & 29.89 & 54.95 & 57.52 & \textbf{69.25} & 45.65 & 50.12 & 74.77 \\
& RCLG   & \textbf{30.17} & \textbf{55.16} & \textbf{59.18} & 68.90 & \textbf{46.16} & \textbf{50.76} & \textbf{75.03} \\
\hline
\multirow{3}{*}{F1} 
& Random & 57.51 & 49.84 & 70.40 & 90.42 & 65.17 & 78.82 & 81.42 \\
& KNN    & 57.68 & 49.15 & 72.38 & \textbf{90.86} & 65.63 & 79.34 & \textbf{83.47} \\
& RCLG   & \textbf{58.12} & \textbf{49.98} & \textbf{72.61} & 90.55 & \textbf{65.92} & \textbf{79.98} & 83.20 \\
\hline
\end{tabular}
}
\caption{Effect of contrastive pair generation strategies (Random vs. KNN vs. RCLG).}
\label{tab:pair_generation}

\end{table}

\subsection{Visualization}
To intuitively assess the impact of information fusion, we visualize clustering results under three model variants: (1) using only the final GNN layer (Single-Layer), (2) integrating multiple layers (Local Fusion), and (3) incorporating global semantics (Local + Global Fusion). All variants share the same backbone and training settings, differing only in the fusion modules, so that the visualization reflects the effect of information integration.

As shown in Figure~\ref{fig:cluster_visualization}, clustering quality improves progressively with each fusion level. In Figure~\ref{fig:cluster_visualization}(a), the DBLP dataset exhibits clustering errors and unclear boundaries when only the final layer is used, while Local Fusion yields more compact clusters by leveraging hierarchical neighborhood cues. In Figure~\ref{fig:cluster_visualization}(b), the AMAP dataset shows more compact and well-separated clusters after incorporating global information, suggesting that semantic prototypes help stabilize representations for nodes in sparse or weakly connected regions. These results demonstrate the effectiveness of combining local and global signals for enhanced embedding quality.